# Semantic Description of Web Services

Thabet Slimani
CS Department, Taif University, P.O.Box 888, 21974, KSA

**Abstract**
The tasks of semantic web service (discovery, selection, composition, and execution) are supposed to enable seamless interoperation between systems, whereby human intervention is kept at a minimum. In the field of Web service description research, the exploitation of descriptions of services through semantics is a better support for the life-cycle of Web services. The large number of developed ontologies, languages of representations, and integrated frameworks supporting the discovery, composition and invocation of services is a good indicator that research in the field of Semantic Web Services (SWS) has been considerably active. We provide in this paper a detailed classification of the approaches and solutions, indicating their core characteristics and objectives required and provide indicators for the interested reader to follow up further insights and details about these solutions and related software.

## 1. Introduction

SWS research has as an objective to combines the services with the aim to achieve given goals. Based on goal descriptions and descriptions of available services, a complex service yielding the desired result is composed automatically. SWS research represents a new line of research on service descriptions and their exploitation. The annotation of services with a description using a formal ontology to express their precise mathematical meaning represents the basic idea of services description in the context of the Semantic Web.

The use of semantics is very useful to enables rich support for handling services. Furthermore, the use of ontologies to annotate services allows a higher degree of automation (describes the services in more formal detail).

The main goal of Semantic Web Services approaches is the automation of service discovery and service composition in a SOA [1].

In the last decade, several approaches have been proposed in the literature and these approaches differ in terms of the formalizations and implementations (Ontology language syntaxes) and in terms of the paradigms proposed for employing these in practice.

This paper is dedicated to provide an overview of these approaches, expressing their classification in terms of commonalities and differences. It provides an understanding of the technical foundation on which they are built. These techniques are classified from a range of research areas including Top-down, Bottom-up and Restful Approaches.

This paper does also provide some grounding that could help the reader perform a more detailed analysis of the different approaches which relies on the required objectives. We provide a little detailed comparison between some approaches because this would require addressing them from the perspective of some tasks supported with Semantic Web Services descriptions (i.e., discovery, invocation, composition, etc) and would also require taking into account the frameworks and developed applications.

The remainder of this paper is organized as follows. Section 2 introduces some principles for Semantic Web Service approaches and present in brief the vast popular of those that have been proposed over the years classified into top-down, bottom-up, and Restful approaches. In Section 3 we provide some information whereupon one could make a more efficient comparison and specified evaluation. This section also provides an organized perspective over the state of the art in Semantic Web Service approaches that can better help understand the evolution of the field. Finally, section 4 provides a conclusion and perspectives for future works.

## 2. Classification of semantic Description of Web Services

The existence of interoperable set of technologies for communication is required for Internet-scale distributed computing. There are currently two major alternative directions in these technologies, named "WS-*" and "REST". The WS-* set of specifications uses the messaging paradigm and specialized service interfaces, with standardized infrastructure protocols (e.g. for security, transactions etc.). The REST direction relies on the architectural style of the World Wide Web and it views Web services as sets of resources accessible through the uniform interface of HTTP. WS-* technologies are mostly deployed within enterprises (and behind firewalls), while the public Web is an increasingly large repository of RESTful services.

Web services in the semantic web are enhanced using rich description languages based on Description Logics (DLs) such as the Web Ontology Language (OWL). However, web services that have been enhanced with formal semantic descriptions is the definition of semantic web services. We distinguish two tested and validated approaches for WS-* technologies in addition to the approach based on REST technologies: Top-Down and Bottom-Up approaches for semantic web services. Top-down approaches are related to the development of semantic web services and are based on the definition of high-level ontologies providing expressive frameworks for describing Web services. On the other hand, bottom-up models, have been adopted an incremental approach that includes semantics to existing Web services standards by adding specific extensions which connects the syntactic definitions to their semantic annotations. Furthermore, the bottom-up approach represents an extension of existing standards and technologies including semantic annotations rather than the entirely services modeling based on ontologies.

If the technical or engineering point of view of a system or an organization seems to be clear and well proved through the history of technology dissemination, then the "top down" strategy is adopted: when all parameters are defined in detailed manner, before implementation, then systems operation works out best. This is the conceptual model for any top down strategy and as application it may be applied to e-government interoperability. As example of e-government application, a powerful administrative organization can be located at the top of hierarchy (e.g. a national government or its agency) and advises the interoperability methods and resources to be applied by all the actors on lower levels, supplements may be made on lower levels respectively.

The bottom up strategy is adopted if everyone concerned bring in his/ her requirements and specifications, and we will find a solution for achieving interoperability within the network which is acceptable for all involved, based on these requirements. For example, if local administrative organizations publish their services interfaces and use his/her individual ontologies, then some joint or mutual service should resolve some technical, syntactic and semantic differences as much as possible. As example of e-government application, administrative organizations can be located at the bottom of the hierarchy which recommend and share interoperability methods and resources from their point of view; and furthermore, the centralized direction is only accepted when there is agreement on all lower levels.

As a Web service domain, we consider both commercial and governmental Web services. A case study based on analysis of 493 commercial and 96 governmental Web service operations has been conducted in the work of Kungas and Matskin., 2006 [2] and the result of the analysis of the interaction between commercial and governmental Web services turned out that while ontologies enhance the usage of the commercial Web services, they have no significant impact on the governmental Web services. However, ontologies facilitate automation of semantic integration of commercial Web services with governmental ones. Based on this analysis, we say that TOP-Down approaches are useful when we faced with commercial Web services use.

Additionally, this idea is confirmed by the work presented in [3] which says that the existence of a web services description in a machine-understandable fashion is expected to have a great impact in areas of e-Commerce and Enterprise Application Integration (EAI).

In the remainder of this section, several languages have been presented and classified.

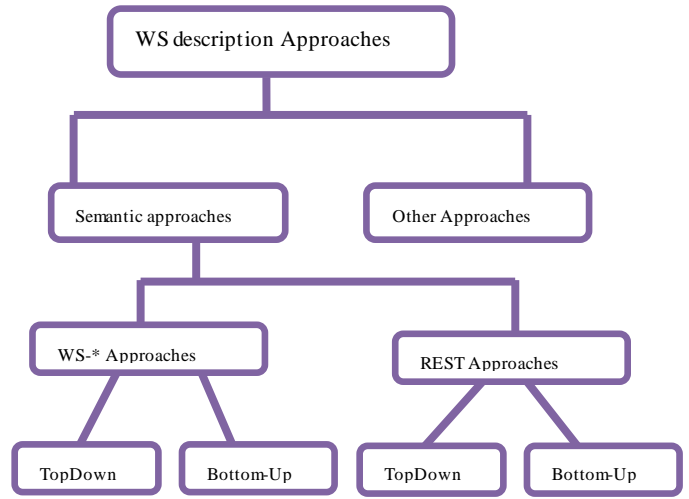

Fig. 1 WS description approaches taxonomy.

## 2.1 Approaches Using WS-* Technologies

### 2.1.1 Top-Down Approaches

The term Top-Down means that semantic web services are written directly in a formal language and don't have any dependence to any non-semantic web services. All semantic web services technologies should be able to connect with non-semantic web services (called grounding) in order to enhance any web service system development. The ability to build new SWS with no relation to the classic web services technologies is the needed features that should characterize this approach. Several languages have been presented for Top-Down approaches:

- **OWL-S** [4] it is mainly a North American development effort, based on the OWL ontology language. The OWL Services (OWL-S) ontology defines an OWL ontology composed by a set of essential vocabularies to describe the "semantics" of Web services. This semantics includes the definitions of the capabilities, requirements, internal structure and interactions details with the service.
- **WSMO** [5] [6] it is a project developed within EU-funded projects (Sekt, DIP, Knowledge Web, ASG and SUPER projects) based on the WSML [6] ontology language. WSMO is a framework for Semantic Web Services that represents a top-down model identifying semantics of web service that uses the WSML (Web Service Modeling Language) language for describing domain-specific semantic models. The description of functional capabilities of services using logical expressions as preconditions, assumptions, postconditions and effects are required by WSMO.
- **SWSL**: SWSL is used to specify the semantics of web services concepts and descriptions as well as individual web services. It includes two sublanguages: *SWSL-FOL* is based on first-order logic (FOL) and is designed primarily to express the formal characterization (ontology) of Web service concepts. *SWSL-Rules* is based on the logic-programming (or "rules") paradigm and is designed to support the actual language for service specification that use the service ontology in reasoning and execution environments based on that paradigm.
- **DIANE**: DIANE is a framework that allows the automation of the discovery, composition, binding and invocation of services [7]. The framework is based on DIANE Service Description (DSD) and a specialized ontology language for describing service elements called DIANE Elements. DIANE elements exploit the notions of attributes, and reuse the clean separation between schema and instances promoted by description logics. Furthermore, special constructs are included in DIANE elements to describe service such as declarative and fuzzy set as well as variables.
- **SWSO**: The Semantic Web Services Ontology (SWSO) is a part of SWSL language [8], which includes formal conceptual definitions and individual web services. The definition of semantics of the theoretic model of the ontology of SWSO is based on the description of the ontology services, and the description of a first-order logic (FOL) axiomatization (FLOWS - the First-order Logic Ontology for Web Services). The aim of the created service descriptions enable automated discovery, composition, and verification, as well as the creation of declarative descriptions of a Web service that can be mapped to executable specifications.
- **COWS**: The Core Ontology of Web Services (COWS) is based on the Core Ontology of Software Components [9]. To enable extensibility and facilitate reuse, the fundamental concepts of COWS are separated in core ontology. The Core Ontology of Software Components is based on fundamental concepts and associations like software, data, users, policies and so on.
- **MSM**: Minimal Service Model (MSM) introduced together with hRESTS [10] is a simple RDF vocabulary covering what can essentially be considered the core of WSDL. It defines basically *Services* characterized by a number of *Operations* which have an *Input*, an *Output*, and *Faults*. Furthermore, MSM has subsequently been used as a means to integrate heterogeneous services (i.e., WSDLs and Web APIs). The combination between MSM and WSMO-Lite can provides a common framework covering the largest common denominator of the most used SWS formalisms on the Web. With this combination generic publication and discovery machinery has been developed that supports SAWSDL, WSMO-Lite, hRESTS/MicroWSMO, and OWL-S services [11].
- **ServONT**: is an ontology-based hybrid approach designed to improve the effectiveness and the efficiency of service discovery. For this matter, additional semantics is associated to the service ontology **ServOnt**, which organizes services at different levels of abstraction by means of semantic relationships that can be fruitfully exploited to support service discovery. Starting from the bottom layer, we distinguish between *Concrete Services*, *Abstract Services* and *Service Categories*, organized into *Concrete*, *Abstract* and *Category* layer, respectively [12].

❖ **SSWAP**: Simple Semantic Web Architecture and Protocol (SSWAP) is the driving technology for the iPlant Semantic Web Program[1]. It combines Web service functionality with an extensible semantic framework to satisfy the conditions for high throughput integration [13]. SSWAP utilizes OWL ontologies to describe the features and capabilities of Web services and standard HTTP methods to invoke the services. The architecture of SSWAP is based on five basic concepts Provider, Resource, Graph, Subject, and Object. The Provider organization is the owner and the publisher of resources. The web pages, ontologies and databases, represent the resources which are used to describe services offered on the Web.

In order to illustrate a scenario of an application that adopts a top-down approach, we will briefly describe an application scenario based on [14]. Let us imagine a "Virtual Traveling Agency" (VTA for short) which is a platform providing eTourism services. These services can cover information services concerned with tourism such as events and sights in different areas and services that support booking of flights, hotels, rental cars, etc. By applying Semantic Web Services, a VTA can invoke Web services provided by several eTourism suppliers and aggregate them into new customer services in a semi-automatic fashion.

2.1.2 Bottom-Up Approaches

The aim of annotating Web Services is to add clarity in the Web Service definitions and also to allow the Web Service to be read by machines. This machine-readability increases the power of the SWS by adding the understanding of what the web Service is doing and the ability to interpret the messages that are interchanged. Semantic annotations of web service are used to automate service discovery, composition, mediation, and monitoring. We can state several approaches actually finished:

➢ **WSDL-S** [15]: WSDL-S specification is a W3C member submission that defines annotations to WSDL documents. The approach of semantic annotation consists in directly annotating the WSDL with semantic information. Semantic annotations that reference concepts in an ontology define the meaning of the inputs, outputs, preconditions and effects of the operations described in a service interface.

➢ **SAWSDL**[2] [16]: SAWSDL is s a W3C proposed recommendation where the semantic annotations use an extended attributes called modelReference so that relationships between WSDL components and concepts in another semantic model (e.g. ontology) are handled. Hence, the separation of semantic annotation mechanism from the representation of the semantic descriptions makes SAWSDL an approach independent of the semantic representation language. As a result, developer's community has more flexibility to select their favorite semantic representation language, to reuse semantic domain models and annotate descriptions using multiple ontologies.

The described approaches present a main advantage of preparing annotation directly in the WSDL XML Schema.

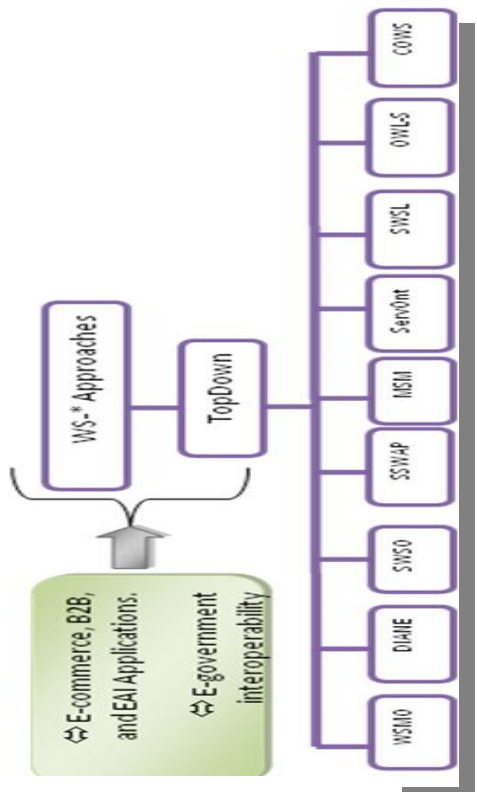

Fig.2 Top-Down WS-* Approaches

---

[1] http://sswap.info

[2] http://www.w3.org/TR/sawsdl/

Other advantage is that these specifications are independent to ontology language. Both languages have the necessary development tools and are operational to model and run SWS.

- ❖ **METEOR-S**: METEOR-S is an effort to create Semantic Web processes, at the LSDIS lab, University of Georgia. METEOR-S is a framework for semi-automatically marking up web service descriptions with ontologies. It contains an algorithms development to match and annotate WSDL files with relevant ontologies. It provides a mechanism to add data, functional and QoS semantics to WSDL files [17].

- ❖ **MWSAF (METEOR-S Web Service Annotation Framework)** is a semantic web based graphical tool that enables you to annotate existing Web service descriptions with ontologies. It facilitates the parsing of WSDL files and ontologies. This enables the user to annotate Web service descriptions semi-automatically. MWSAF was formerly known as **SAWS (Semantic Annotation of Web Services)**. MWSAF offers various features for programmers looking to create Semantic Web services. It provides: a) a fast and easy method for annotating WSDL files with single or multiple ontologies, b) an intuitive graphical environment for viewing WSDL files as well as ontologies, c) support for RDF-S , DAML+OIL and OWL based ontologies and d) a good solution for selecting the correct domain ontology for annotation from several ontologies.

- ❖ **USDL:** Universal Service-Semantic Description Language "is a language for formally describing the semantics of Web services" [18]. The USDL common basis that understands the meaning of services is based on OWL and the use of WordNet. The first attempt of USDL is to capture the semantics of web-services in a universal, yet decidable manner [18]. *USDL is designed* based on two languages: WSDL and OWL and defines a generic class called Concept, which is used to define the semantics of messages parts. The *USDL* Concept class denotes the conceptual objects constructed from the OWL WordNet ontology.

- ❖ **ServFace**: The ServFace project [19] aims at creating a model-driven service engineering methodology for an integrated development process for service-based applications[1]. The aims of this approach is to add UI-related annotations to service descriptions, notably WSDLs, in order to better support the development of user interface and to build interactive service-based applications. This project includes the creation of new algorithms for the composition of annotated services to build interactive service based applications based on the user interface annotations.

- ❖ **GPO/PSAM**: The General Process Ontology (GPO) and the Process Semantic Annotation Model (PSAM) [20] define business process annotations. The GPO/PSAM approach has been developed into a complete and systematic semantic annotation framework and defines four perspectives: basic description of process models (profile annotation), process modeling languages (meta-model annotation), process models (model annotation) and the purpose of the process models (goal annotation). Profile annotations are basic process description and include the following groups: administrative (e.g., creator, publisher), descriptive (e.g., title, category), technical (modeling language), preservation (documentation) and use (e.g., used in). Meta-model annotations include typical business process constructs such as: **Activity, Actorrole, Input, Output, Merge, Join,** and others. Model annotations use process modeling ontology as metadata to annotate the semantics of constructs in a modeling language. Goal annotations are used to specify aims of business process activities with distinction on *local* and *global* goals.

- ❖ **QuASAR [2] / ISPIDER:** The goal of Quality Assurance of Semantic Annotations for Services (QuASAR) [21] is to support the full life-cycle of Web service annotations and to ensure trustworthiness and accuracy of annotations. QuASAR / ISPIDER approach explores the potential uses of an additional source of information about semantic annotations: namely, repositories of trusted data-driven workflows. A workflow is a network of service operations, connected together by data links describing how the outputs of the operations are to be fed into the inputs of others. If a workflow is known to generate sensible results, then it must be the case that the operation parameters that are connected within the workflow are compatible with one another (to some degree). Semantic annotations have been proposed as a means of providing richer information about the behavior of Web services to potential users [21]. Three proposed ontologies of terms used in of service annotation[1]: *Domain ontology*, *Representation ontology* and *Extend ontology*. Domain ontology represents service annotations from similar a domain (e.g., biomedical services and others) that describes common concepts relevant within a given domain. The description of the representation format of service parameters is obtained by the Representation ontology. Extend ontology describes scopes of values of service parameters. Information about scopes of values helps to detect incompatibilities between well formed services.

---

[1] http://www.servface.eu/

[2] http://img.cs.manchester.ac.uk/quasar/

- ❖ **BPEL4SWS** :

  BPEL4SWS [22] is a language for Semantic Web Service orchestration based on Business Process Execution Language (BPEL). BPEL is an orchestration language that defines business processes interacting with external entities through web service operations using WSDL. BPEL4SWS extends BPEL and enables the definition of process logic independently from WSDL specific details. It is useful for orchestration of both Web services and Semantic Web Services. Semantic annotations can be attached to any part of BPEL4SWS descriptions. It allows the functionality descriptions or requirements of activities of a process semantically using SWS frameworks such as WSMO or OWL-S instead of using WSDL. BPEL4SWS also makes use of the SAWSDL standard for handling data lifting and lowering and enables bridging the gap between XML data and ontologies and enables semantic service discovery using appropriate middleware such as SEE during runtime.

- ❖ **YASA4WSDL**: Yet Another Semantic Annotation (YASA) for WSDL [23] proposes an extension of SAWSDL. YASA4WSDL includes two types of ontologies: The first one is a *Technical Ontology* containing concepts for ontologies describing service concepts (interface, input, output) and ontologies describing non functional concepts of services (ex. QOS attributes). The second type is a *Domain Ontology* that covers a business domain. YASA claims that introducing *serviceConcept* attribute makes SAWSDL descriptions more expressive and allows to explicitly capturing information on service pre-, post-conditions and effects. The separation of semantic annotation mechanism from the representation of the semantic descriptions makes SAWSDL an approach independent of the used semantic representation language.

- ❖ **WSMO-Lite:** Has been created due to a need for lightweight service ontology which would directly build on the newest W3C standards and allow bottom-up modeling of services. WSMO-Lite adopts the WSMO model and makes its semantics lighter and allows the use of any ontology language with RDF syntax. WSMO-Lite only defines semantics for the information model, functional and nonfunctional descriptions (as WSMO Service does) and only implicit behavior semantics.

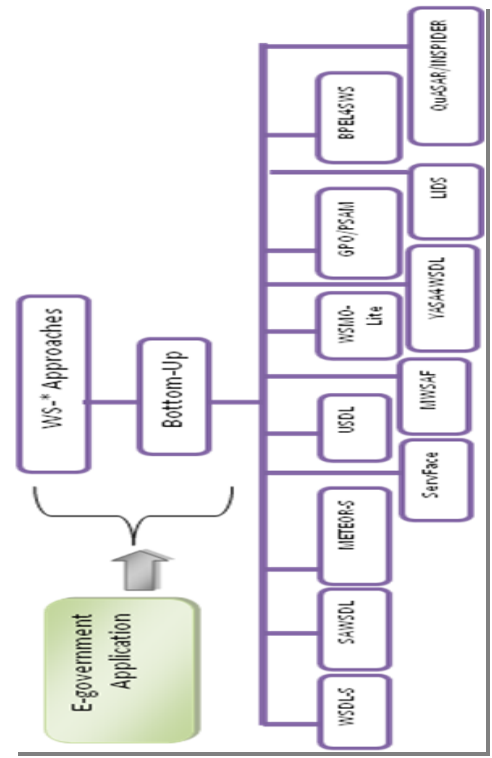

Fig. 3 WS-* Bottom-up Approaches Taxonomy.

---
[1] http://img.cs.manchester.ac.uk/quasar/

- **LIDS:** Linked Data Services (LIDS) [24] denote the integration of data providing services and linked data and represents a lightweight service description model where service inputs and outputs are specified using SPARQL graph patterns. It focuses on the integration of existing data services exposed with Linked Data principles through Web APIs. Furthermore, the Web standards such as HTTP, RDF and SPARQL represent the base of LIDS. In addition to its accessibility over HTTP protocol, LIDS consume and produce RDF triples. LIDS can be directly used by Linked Data consumers and any requirement for data lifting.

2.2 Approaches Using REST Technologies

2.2.1 Top-Down Approaches

RESTful services are currently facing similar limitations to those identified for traditional Web service technologies and present even further difficulties, such as the lack of machine-processable service descriptions. Traditional Web service technologies have a somewhat longer history of research on semantic descriptions and annotation approaches; research in the area of semantic RESTful services is newer and therefore relatively limited. In order to address these challenges and to enable the wider adoption of RESTful service technologie, the following approaches have been developed.

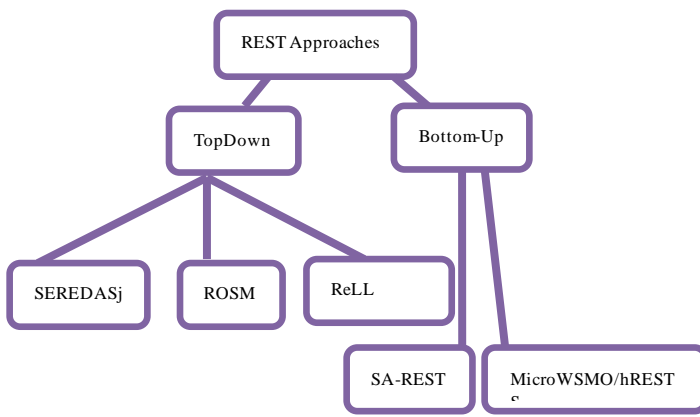

Fig. 4 **RESTful** Approaches Taxonomy.

- **ROSM:** The Resource-Oriented **Service Model (ROSM)**[1] ontology is a lightweight approach to the structural description of resource-oriented (RESTful) services. The use of ROSM enables the annotation of resources included in a service. Furthermore, a resource can be described as a part of collections and accompanied with addresses (URIs) intended for access and manipulation. A resource can be organized in collections, allowing the capture of an arbitrary number of resources and attaching service semantics to them following the SAWSDL approach.

- **SEREDASj**: stands for SEmantic REstful DAta Services, while the "j" should high-light that the approach is based on JSON (this leaves the door open for other data formats). SEREDASj semantically describe RESTful Data Services which in consequence leads to a mechanism to transform the data provided by such services to semantic resources. This aims to contribute to the availability of more semantic datasets [25].

- **Rell**: ReLL [26], the Resource Linking Language, does exactly the opposite. ReLL is a language to describe RESTful services with the aim to transform their exposed data to RDF and thus allowing harvesting already existing Web resources. Currently ReLL does not support any modification of the described re-sources, i.e., at the moment it supports only HTTP GET operations. This clearly restricts the possible use cases of ReLL at this point in time.

2.2.2 Bottom-Up Approaches

We consider the following bottom-up approaches:

- **MicroWSMO/hRESTS**:MicroWSMO [10] is a formalism for the semantic description of RESTful services, which is based on adapting the SAWSDL approach that adds sematic annotations. MicroWSMO uses microformats for adding semantic annotation ro service properties on top of HTML service documentation, by relying on hRESTS (HTML for RESTful Services) (Kopecky et al. 2008) that introduces the service model structure (service, operations, input, output) that allows the descriptions machine-processable. hRESTS enables the annotation of service operations, inputs and outputs, HTTP methods and labels, by inserting HTML tags within the HTML. MicroWSMO enables the identification of RESTful services and brings them to a level where they can be more easily discovered, composed and invoked.

---

[1] F. F. and N. B. D3.4.6 MicroWSMO v2 – Defining the second version of MicroWSMO as a systematic approach for rich tagging. Soa4all project deliverable.

- ❖ **SA-REST**: Semantic Annotations for REST (SA-REST) [27] is an open, standards-based approach which adds semantic annotations to RESTful services and Web APIs [15]. SA-REST defines three basic properties that can be used to non-intrusively annotate HTML/XHTML documents, typically to embed ontological meta-data[1]: The *domain-rel* property that provide domain information descriptions for a resource. The main objective of this annotation is to provide coarse grained categorizations of the HTML elements. The *sem-rel* property, which refers to the popular rel tag, and used to capture the semantics of a link within an HTML document. This kind of annotation is supposed to be used only within an anchor element (<a>). Finally, the *sem-class* property can be used to single entity annotation within a resource.

## 3. Comparison of SWS Approaches Functionalities

In the previous section we have briefly introduced the different approaches proposed in the literature, providing a basic description and pointers for the interested reader. In this section, we provide a comparison between these models in terms of their goal of development, their representation language, their conceptual influences and the year they were proposed in. This comparison is located in the Tab 1.

The lack of freely offered services and the acquisition of service descriptions, or the complexity of this task is the major limitation of the efforts described before. Some recent efforts aim to resolve this problem by reducing the complexity of the models and the acquisition task, by using simple RDF(S) vocabularies and Linked Data. These recent approaches present some promising results that could certainly be beneficial for the SWS paradigm.

OWL-S and WSMO is fully edged semantic framework, but WSDL and SAWSDL lack the support for semantic description. The most mature and commonly used in service discovery and composition approaches is OWL-S. But OWL-S presents some drawbacks as stated in [28]:
The process model of OWL-S is neither an orchestration model nor a choreography model. Moreover, OWL-S views Web service description does not consider asynchronous communication, and take into account only synchronous communication. The process model of WSMO offers both an orchestration and choreography view, but the orchestration view is rather primitive and the WSMO Choreography model contains transition rules which represent only local constraints. Furthermore, WSMO hasn't been around as long as OWL-S.

None of the approaches described in the Table 1 provide a complete solution according to the dimensions illustrated, but interestingly WSMO shows complementary strengths because it allows several goals (Discovery, Composition, Invocation, Orchestration and Mediation) and partially SWSO which not allows only the Mediation process.

Additionally the characteristic of UPML, OWL-S, DIANE, GPO and SWSO is very interesting being given that they allows functional, non-functional, informational and behavioral descriptions.

## 4. Conclusions

Research on SWS has produced several conceptual models, languages, architectures and algorithms that express the potential of these technologies for the Web and organizations. In this paper we have provided an initial description of these works according a number of dimensions. This paper is a first step that presents a breadth of the field, principally in terms of the tasks that could be supported by means of SWS descriptions, allowing a good state of the art and a comparison in the field. The use of SWS on the Web is unusual and it looks like that the intelligent techniques of the Web that act to the users profit remains as indicated by the reputation of publicly available Web APIS and RESTful services.

It is required to use the domain ontologies, the services taxonomies and in some cases to include complicated logical expressions, in order to create a rich semantic description of a Web service.

---

[1] http://www.w3.org/Submission/2010/SUBM-SA-REST-20100405/

**Thabet Slimani** graduated at the University of Tunis (Tunisia Republic) and defended PhD. thesis with title "New approaches for semantic Association Extraction and Analysis". He has been working as an assistant Professor at the Department of Computer Science, Taif University. He is a member of Larodec Lab (Tunis University). His interests include semantic Web, data mining and web service. He is author of more than 20 scientific publications.

Tab. 1 Comparison between SWS approaches functionalities.

| Approach | Year | Conceptual Input | Language | Goal | | | | |
|---|---|---|---|---|---|---|---|---|
| | | | | Discovery | Composition | Invocation | Orchestration | Mediation |
| **UPML** | 1999 | PSMs | UPML/LISP | Knowledge-Based Systems development | | | | |
| **DAML-S /OWL-S** | 2001 | Agents | Knowledge-Based Systems development | Semantic annotations of WS | | | | |
| **DIANE** | 2004 | OWL-S | DIANE Elements | √ | √ | √ | x | x |
| **SWSO** | 2005 | OWL-S | SWSL | √ | √ | √ | √ | x |
| **USDL** | 2005 | OWL-S | OWL | Language for formally describing the semantics of Web services | | | | |
| **WSDL-S** | 2005 | WSMO, OWL-S, WSDL | XML Shema | Linking semantic annotations to Web services | | | | |
| **WSMO** | 2005 | WSMF, UPML | WSML, RDF | √ | √ | √ | √ | √ |
| **COWS** | 2006 | DOLCE | OWL | Semantic management of middleware | | | | |
| **GPO** | 2006 | UEMO | OWL | Process modeling | | | | |
| **QuASAR** | 2006 | ᵐʸGrid | OWL | Integrated platform enabled as Grid and Web services for the storage, dissemination and management of proteomic data | | | | |
| **WSO** | 2006 | OWL-S, WSMO, WSBPEL | OWL | x | √ | x | x | x |
| **BPEL4SWS** | 2007 | BPEL4WS, WSMO, SAWSDL | XML Shema | x | x | x | √ | x |
| **SAWSDL** | 2007 | WSDL-S | XML Schema | Semantic Annotations for WS WSDL and XML Schema | | | | |
| **FUSION Ontology** | 2008 | SAWSDL, UDDI | OWL-DL | Service registry | | | | |
| **YASA** | 2008 | SAWSDL | XML Schema | Extension of SAWSDL, service discovery | | | | |
| **MicroWSMO/ hRESTS** | 2008 | hRESTS/WSMOLite | HTML with microformat tags | Semantic annotations of RESTful services and Web APIs | | | | |
| **MSM** | 2008 | WSDL, WSMOLite, hRESTS | RDF(S) | √ | x | √ | x | x |
| **ServONT** | 2008 | OWL | OWL-DL | √ | x | x | x | x |
| **WSMO-Lite** | 2008 | SAWSDL, OWL-S, WSMO | RDF(S) | √ | √ | √ | x | x |
| **ServFace** | 2009 | WSDL | XML Schema | For adding of UI-related Annotations to Web service Descriptions (WSDL) | | | | |
| **SSWAP** | 2009 | HTML, Semantic MOBY | OWL | Data and service integration in Biology | | | | |
| **SA-REST** | 2010 | SAWSDL, hRESTS | RDFa | Semantic annotations of RESTful services | | | | |
| **ER Model** | 2010 | ER, BPEL | ER, OWL DL | √ | √ | x | x | x |
| **LIDS** | 2010 | HTTP, Linked Data | SPARQL | Bridging the gap between data services and Linked Data principles. Lightweight composition | | | | |
| **RELL** | 2010 | REST | RDF / OWL | Description of resource-centered Web APIs in terms of resources | | | | |
| **ROSM** | 2010 | WSMO-Lite, REST | RDFS, SPARQL | Description of resource-centered Web APIs (RESTful services) | | | | |
| **SEREDASj** | 2011 | JSON-LD, REST | JSON, RDF, FOAF | Semantic description of Restful Data Services | | | | |